\documentclass[preprint,12pt]{elsarticle}

\usepackage{amssymb}
\usepackage{amsmath}

\usepackage{xcolor}
\usepackage{amstext}
\usepackage{multirow}
\usepackage{booktabs}
\usepackage{makecell}
\usepackage{comment}
\usepackage{hyperref}
\usepackage{graphicx}

\journal{The Journal of Biomedical Informatics}

\begin{document}

\begin{frontmatter}



\title{PMC-Patients: A Large-scale Dataset of Patient Summaries and Relations for Benchmarking Retrieval-based Clinical Decision Support Systems}


\author[label1]{Zhengyun Zhao\fnref{eqa}}
\author[label2]{Qiao Jin\fnref{eqa}}
\author[label2]{Fangyuan Chen}
\author[label3]{Tuorui Peng}
\author[label1]{Sheng Yu\corref{cor1}}
\fntext[eqa]{Equal contributions.}
\cortext[cor1]{Corresponding author at: Weiqinglou 209, Center for Statistical Science, Tsinghua University, Beijing, China. Email address: \url{syu@tsinghua.edu.cn}}

\affiliation[label1]{organization={Center for Statistical Science, Tsinghua University},
            addressline={}, 
            city={Beijing},
            postcode={100084}, 
            state={},
            country={China}}

\affiliation[label2]{organization={School of Medicine, Tsinghua University},
            addressline={}, 
            city={Beijing},
            postcode={100084}, 
            state={},
            country={China}}

\affiliation[label3]{organization={Department of Physics, Tsinghua University},
            addressline={}, 
            city={Beijing},
            postcode={100084}, 
            state={},
            country={China}}

\begin{abstract}
\textbf{Objective:}
Retrieval-based Clinical Decision Support (ReCDS) can aid clinical workflow by providing relevant literature and similar patients for a given patient. 
However, the development of ReCDS systems has been severely obstructed by the lack of diverse patient collections and publicly available large-scale patient-level annotation datasets. 
In this paper, we aim to define and benchmark two ReCDS tasks: Patient-to-Article Retrieval (ReCDS-PAR) and Patient-to-Patient Retrieval (ReCDS-PPR) using a novel dataset called PMC-Patients.

\noindent
\textbf{Methods:}
We extract patient summaries from PubMed Central articles using simple heuristics and utilize the PubMed citation graph to define patient-article relevance and patient-patient similarity.
We also implement and evaluate several ReCDS systems on the PMC-Patients benchmarks, including sparse retrievers, dense retrievers, and nearest neighbor retrievers.
We conduct several case studies to show the clinical utility of PMC-Patients.

\noindent\textbf{Results:}
PMC-Patients contains 167k patient summaries with 3.1M patient-article relevance annotations and 293k patient-patient similarity annotations, which is the largest-scale resource for ReCDS and also one of the largest patient collections.
Human evaluation and analysis show that PMC-Patients is a diverse dataset with high-quality annotations.
The evaluation of various ReCDS systems shows that the PMC-Patients benchmark is challenging and calls for further research: the best baseline retriever achieves only 16\% P@10, 63\% R@1k on ReCDS-PAR, and 6\% P@10,  80\% R@1k on ReCDS-PPR.

\noindent\textbf{Conclusion:}
We present PMC-Patients, a large-scale, diverse, and publicly available patient summary dataset with the largest-scale patient-level relation annotations.
Based on PMC-Patients, we formally define two benchmark tasks for ReCDS systems and evaluate various existing retrieval methods.
PMC-Patients can largely facilitate methodology research on ReCDS systems and shows real-world clinical utility.
We release all code and data at \url{https://github.com/pmc-patients/pmc-patients} to benefit the community.

\end{abstract}

\begin{graphicalabstract}
 \begin{figure}[ht!]
    \centering
    \includegraphics[width=\linewidth]{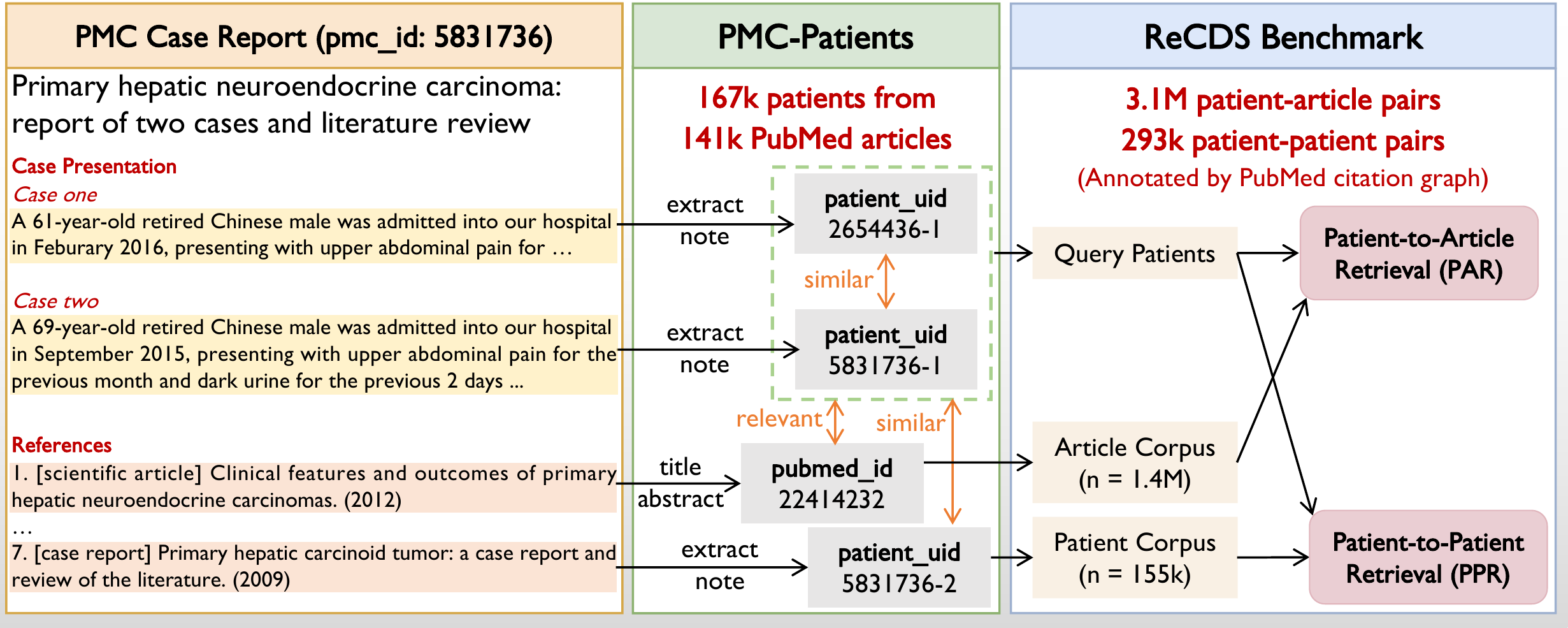}
\end{figure}
\end{graphicalabstract}

\begin{highlights}
\item We introduce PMC-Patients, a first-of-its-kind dataset consisting of 167k patient summaries extracted from case reports in PubMed Central with high-quality annotations.

\item Based on PMC-Patients, we establish two large-scale benchmarks to evaluate retrieval-based clinical decision support (ReCDS) systems: a patient-to-article retrieval task with 3.1M relevant articles and a patient-to-patient retrieval task with 293k similar patients.

\item We systematically evaluate various baseline methods on the proposed ReCDS benchmarks and conduct several case studies to demonstrate the clinical utility. 
\end{highlights}

\begin{keyword}
Dataset collection \sep case report \sep clinical decision support \sep patient similarity \sep information retrieval


\end{keyword}

\end{frontmatter}

\section{Introduction}
\label{intro}
Clinicians often rely on Evidence-Based Medicine (EBM) to combine clinical experience with high-quality scientific research to make decisions for patients \citep{sackett1997evidence}. 
However, finding relevant research can be challenging since the number of scientific publications is growing exponentially, leaving many clinical questions unanswered \citep{ely2005answering}.
To address this issue, there has been increasing research interest in utilizing Natural Language Processing (NLP) and Information Retrieval (IR) techniques to retrieve relevant articles or similar patients for assisting patient management \citep{roberts2016overview, Pan2019AnAT, Park2020AutomaticIO, Zhang2021AnIB, Zhang2023AHA}. 
In this article, we introduce the term ``Retrieval-based Clinical Decision Support" (ReCDS) to describe these tasks. ReCDS can provide clinical assistance for a given patient by retrieving and analyzing relevant articles or similar patients to determine the most likely diagnosis and the most effective treatment plan.

ReCDS with relevant articles is grounded in EBM, where the target articles to retrieve are up-to-date clinical guidelines or high-quality evidence such as systematic reviews.
Therefore, the majority of ReCDS studies have focused on retrieving relevant research articles \citep{Gurulingappa2016SemiSupervisedIR, Sankhavara2018BiomedicalDR, Shi2022HybridRF}, which are primarily facilitated by the Clinical Decision Support (CDS) Track \citep{simpson2014overview, roberts2015overview, roberts2016overview} held annually from 2014 to 2016 at the Text REtrieval Conference (TREC).
Each year, the TREC CDS Track releases 30 ``medical case narratives", which serve as idealized representations of actual medical records, including patient information such as past medical histories and current symptoms.
Participants are asked to return relevant PubMed Central (PMC) articles for each patient with regard to a given aspect (diagnosis, test, or treatment).
Although sufficient article relevance can be annotated for each patient under the TREC pooling evaluation setting \citep{DBLP:conf/sigir/BuckleyV04}, the size and diversity of the test patient set in TREC CDS are very limited.
Consequently, the generalizability of system performance to uncovered medical conditions may be constrained.

ReCDS with similar patients, on the other hand, is much under-explored.
In brief, ``similar patients with similar features have similar outcomes'' \citep{seligson2020recommendations}. 
Retrieving the medical records of similar patients can provide valuable guidance, especially for patients with uncommon conditions such as rare diseases that lack clinical consensus.
Nevertheless, there are various challenges in conducting this type of research.
Unlike scientific articles, there is currently no publicly available collection of ``reference patients" to retrieve from.
Moreover, defining ``patient similarity" is non-trial \citep{seligson2020recommendations} and large-scale annotation is prohibitively expensive. 
As a result, there are only a few studies on similar patient retrieval \citep{plaza2010retrieval, arnold2010clinical}, all of which use private datasets and annotations.

The aforementioned issues make it clear that a standardized benchmark for evaluating ReCDS systems is greatly needed.
Ideally, such a benchmark should contain:
(1) a diverse set of patient summaries, which serve as both the query patient set and the reference patient collection;
(2) abundant annotations of the patient summaries with relevant articles and similar patients.
Due to privacy concerns, only a few clinical note datasets from Electronic Health Records (EHRs) are publicly available.
One notable large-scale public EHR dataset is MIMIC \citep{johnson2016mimic, mimiciv}. 
However, it only contains ICU patients without any relational annotations, making it unsuitable for evaluating ReCDS systems.

In this article, we aim to benchmark the ReCDS task with PMC-Patients, a novel dataset collected from the case reports in PMC and the citation graph of PubMed.
Case reports denote a class of medical publication that typically consists of: 
(1) a case summary that describes the patient's admission, treatment, progress, discharge, and follow-up situations; 
(2) a literature review that discusses similar cases and relevant articles, which are cited and recorded in the PubMed citation graph.
To build PMC-Patients, we first extract 167k patient summaries from case reports published in PMC using simple heuristics.
For these patient summaries, we then annotate 3.1M relevant articles and 293k similar patients using the PubMed citation graph.
PMC-Patients is one of the largest patient summary collections, with the largest scale of relation annotations for benchmarking ReCDS.
Besides, the patients in our dataset show a much higher level of diversity in terms of demographics and medical conditions than existing patient collections.
Our manual evaluation shows that both patient summaries and relation annotations in PMC-Patients are of high quality. 

Based on PMC-Patients, we formally define two ReCDS tasks: Patient-to-Article Retrieval (ReCDS-PAR) and Patient-to-Patient Retrieval (ReCDS-PPR).
We systematically evaluate the performance of various feature-based and learning-based ReCDS systems, and the experimental results show that both ReCDS-PAR and ReCDS-PPR are challenging tasks and we call for further improvements.
We also present highly-relevant case studies to demonstrate the potential application and significance of our retrieval tasks in three typical clinical scenarios.

\begin{figure}[ht!]
    \centering
    \includegraphics[width=\linewidth]{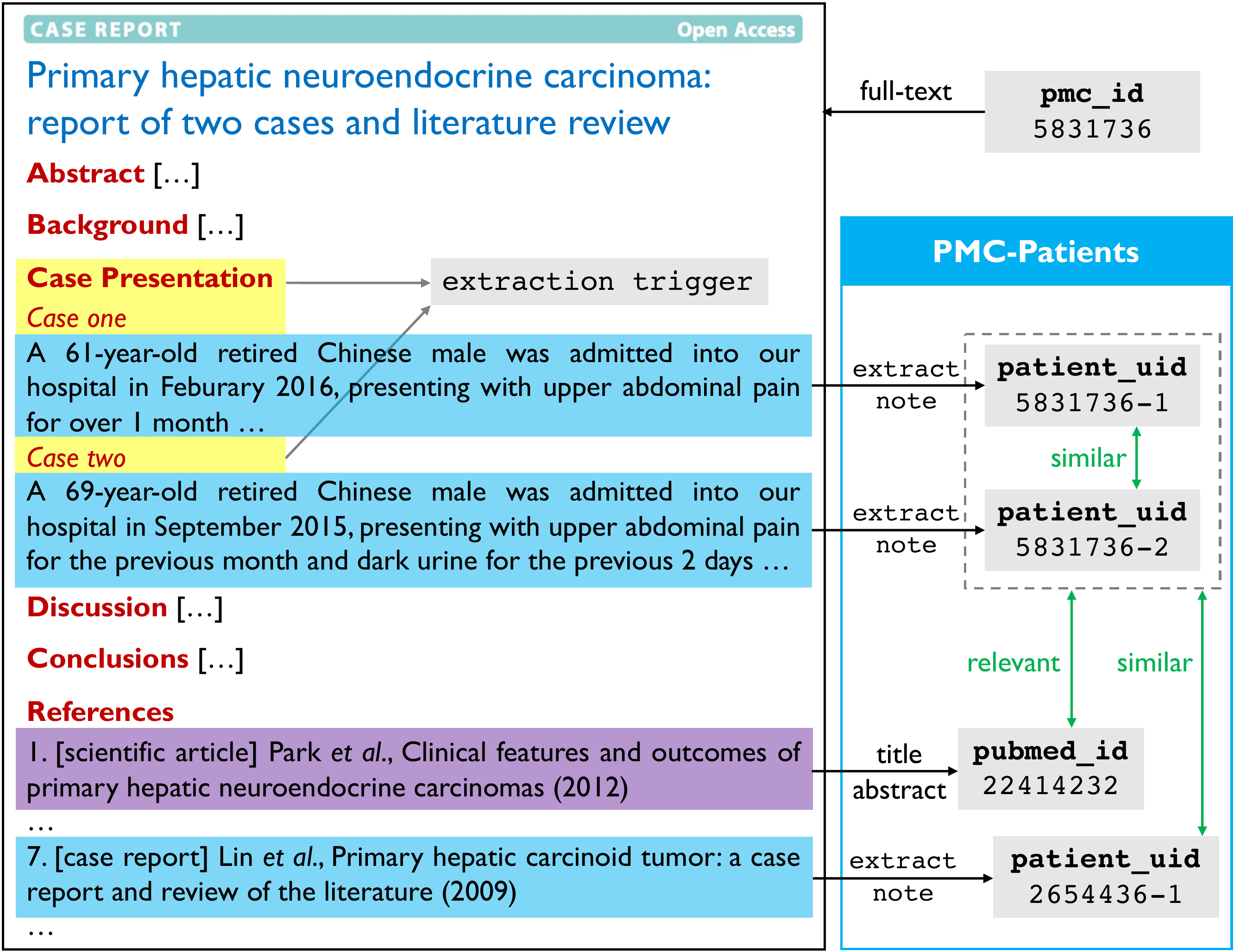}
    \caption{Overview of the PMC-Patients dataset architecture. Patient summaries are extracted by identifying certain sections in PMC articles. The cited articles and patients are considered relevant and similar, respectively. Patients from the same report are also considered similar.}
    \label{fig:overview}
\end{figure}

\begin{figure}[ht!]
    \centering
    \includegraphics[width=\linewidth]{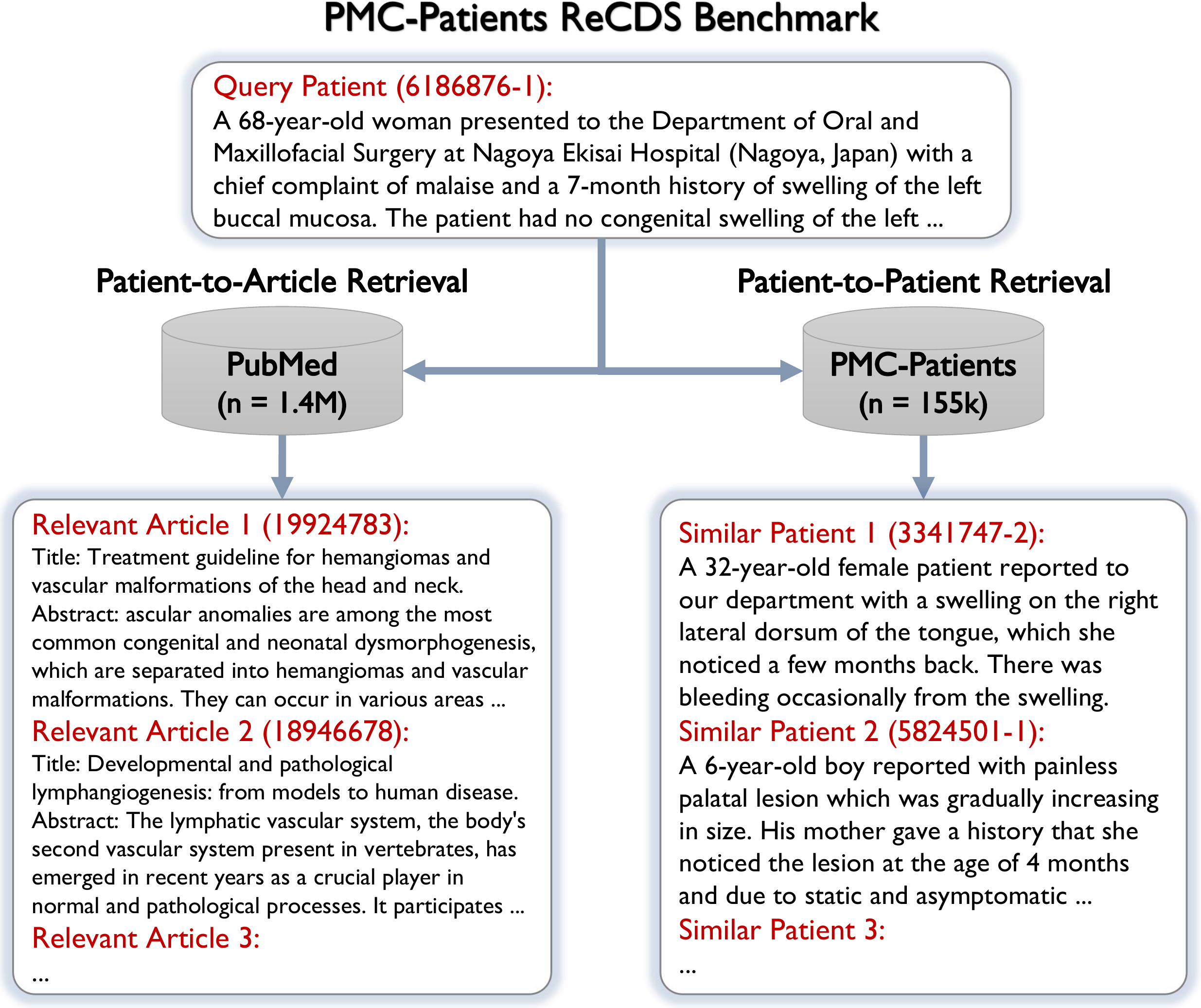}
    \caption{Overview of the PMC-Patients ReCDS benchmark. Given a query patient, there are two tasks: 1. Patient-to-article retrieval requires returning relevant articles from PubMed; 2. Patient-to-patient retrieval requires returning similar patients from PMC-Patients.}
    \label{fig:ReCDS}
\end{figure}

Figure \ref{fig:overview} and Figure \ref{fig:ReCDS} show an overview of our dataset and ReCDS benchmark, respectively.
In summary, the key contributions of this study are three-fold:
\begin{enumerate}
    \item We introduce PMC-Patients\footnote{Available at \url{https://github.com/pmc-patients/pmc-patients}}, a first-of-its-kind dataset consisting of 167k patient summaries extracted from case reports. We systematically characterize PMC-Patients, and show that it is a diverse dataset with high-quality annotations.
    \item Based on PMC-Patients, we formally define two tasks and provide the largest-scale resources to benchmark Retrieval-based Clinical Decision Support (ReCDS) systems: Patient-to-Article Retrieval (ReCDS-PAR) and Patient-to-Patient Retrieval (ReCDS-PPR). ReCDS-PAR contains 3.1M relevant patient-article pairs, and ReCDS-PPR contains 293k similar patient-patient pairs.
    \item We systematically evaluate various ReCDS systems on the PMC-Patients benchmark. We also conduct several case studies to demonstrate the clinical utility of PMC-Patients. 
\end{enumerate}

\section{Material and methods}\label{method}
To collect the PMC-Patients dataset, we utilize the full-text literature resources in PubMed Central (PMC)\footnote{\url{https://www.ncbi.nlm.nih.gov/pmc/}} and the citation relationships in PubMed\footnote{\url{https://pubmed.ncbi.nlm.nih.gov/}}, which will be described in Section \ref{collection}.
Based on PMC-Patients, we formally define two ReCDS benchmarks: Patient-to-Article Retrieval (ReCDS-PAR) and Patient-to-Patient Retrieval (ReCDS-PPR).
We will present the tasks in Section \ref{task_def}, and introduce the baseline methods in Section \ref{baseline}.

\subsection{PMC-Patients Dataset} \label{collection}

We only use PMC articles with at least CC BY-NC-SA license (about $3.2$M) to build the redistributable PMC-Patients dataset.
The collection pipeline can be summarized in three steps (the implementation details and graphical illustration of the pipeline can be found in \ref{extraction_module}):

\begin{enumerate}[(a)]
    \item We identify potential patient summaries in each article section using \textbf{extraction triggers}, which are a set of regular expressions searching for specific patterns of patient summaries, such as ``Case report'' and ``Patient representation'' in the section title.
    \item For sections identified in (a), we extract patient summary candidates using several \textbf{extractors}. Extractors operate at the paragraph level, so a candidate patient summary always consists of one or several complete paragraphs. 
    Besides, we also extract the candidates' demographics (ages and genders) using regular expressions.
    \item We apply various \textbf{filters} to each candidate patient summary extracted in (b) to exclude candidates that are too short, non-English, or without patient demographics.
\end{enumerate}

For each extracted patient summary in PMC-Patients,
we use the citation graph of PubMed\footnote{Extracted from the PubMed baseline updated until July 2022 (\url{https://ftp.ncbi.nlm.nih.gov/pubmed/baseline/}).}
to automatically annotate (1) \textit{relevant articles} in PubMed and (2) \textit{similar patients} in PMC-Patients. 

\textbf{Annotating relevant articles}: 
We assume that if a PubMed article cites or is cited by a patient-containing article, the article is relevant to the patient.
Formally, we denote a patient as $p$, and the article that contains $p$ as $a(p)$.
We define any PubMed article $a'$ relevant to the patient $p$, denoted as $Rel(p, a') = 1$, if: 
$a' \xrightarrow{\text{cites}} a(p)$, 
or $a(p) \xrightarrow{\text{cites}} a'$, 
or $a(p) = a'$ (to ensure label completeness).

\textbf{Annotating similar patients}:
We annotate similar patients based on relevant articles.
For each patient in PMC-Patients, if its relevant articles contain other patients in the dataset, we will label them as similar patients.
Formally, we define any two patients $p_x$ and $p_y$ in PMC-Patients similar, denoted as $Sim(p_x, p_y) = 1$, if: 
$a(p_x) \xrightarrow{\text{cites}} a(p_y)$,
or $a(p_y) \xrightarrow{\text{cites}} a(p_x)$,
or $a(p_x) = a(p_y)$.

\subsection{PMC-Patients ReCDS Benchmarks} \label{task_def}
The PMC-Patients dataset contains $167$k patient summaries, annotated with $3.1$M relevant articles and $293$k similar patients. 
Based on the dataset, we define two benchmarking tasks for ReCDS: Patient-to-Article Retrieval (ReCDS-PAR) and Patient-to-Patient Retrieval (ReCDS-PPR). 
Both are modeled as information retrieval tasks where the input is a patient summary $p \in \mathcal{P}$, where $\mathcal{P}$ denotes the PMC-Patients dataset. 
For ReCDS-PAR, the objective is to retrieve PubMed articles relevant to the input patient from the corpus $\mathcal{A}$. 
Instead of using the entire $33.4$M articles in PubMed, we restrict the retrieval corpus to contain only articles relevant to at least one patient.
Formally, $\mathcal{A} = \{a \ |\  \exists p \in \mathcal{P}, Rel(p, a) = 1 \}$ and contains 1.4M articles, which is a more feasible setting.
For ReCDS-PPR, the objective is to retrieve patients similar to the input patient from PMC-Patients.
The benchmark statistics are shown in Table \ref{tab:benchmark_stat}.

\begin{table}[ht!]
    \small
    \centering
        \begin{tabular}{lccccccc}
        \toprule
        \multirow{2}{*}[-1.5ex]{\textbf{Split}} & \multirow{2}{*}[-1.5ex]{\textbf{\thead{Source\\Articles}}} &
        \multicolumn{3}{c}{\textbf{ReCDS-PAR}} & \multicolumn{3}{c}{\textbf{ReCDS-PPR}} \\
        \cmidrule(r){3-5} \cmidrule(r){6-8} & & \textbf{\thead{Query\\Patients}} & \textbf{\thead{Relevant\\Articles}} & \textbf{A/P} & \textbf{\thead{Query\\Patients}} & \textbf{\thead{Similar\\Patients}} & \textbf{P/P} \\
        \midrule
        train & 131k & 155.2k & 2.9M & 18.64 & 94.6k & 257.4k & 2.72\\
        dev & 5k & 5.9k & 107.5k & 18.14 & 2.9k & 6.4k & 2.22\\
        test & 5k & 6.0k & 114.1k & 19.1 & 2.8k & 7.5k & 2.66\\
        \midrule
        \multicolumn{2}{c}{\textbf{Corpus}} & \multicolumn{3}{c}{1.4M candidate articles} & \multicolumn{3}{c}{155.2k candidate patients} \\
        \bottomrule
        \end{tabular}
    \caption{Statistics of the ReCDS-PAR and ReCDS-PPR benchmarks. 
    A/P: Average number of relevant articles per query patient.
    P/P: Average number of similar patients per query patient.
    }
    \label{tab:benchmark_stat}
\end{table}

We split the train/dev/test on the article level. 
Specifically, we randomly select two subsets of articles (5k in each) from which PMC-Patients is extracted and include the corresponding patients in the dev and test dataset as query patients. 
Patient summaries extracted from other articles are included as the training query patients and also used as the retrieval corpus ${\mathcal{P}}$.

We evaluate retrieval models on both benchmarks with Mean Reciprocal Rank (MRR), Precision at 10 (P@10), normalized Discounted Cumulative Gain at 10 (nDCG@10), and Recall at 1k (R@1k).

\subsection{Baseline models}\label{baseline}
We implement three types of baseline retrieval models for both ReCDS-PAR and ReCDS-PPR: sparse retriever, dense retriever, and nearest neighbor retriever.

\textbf{Sparse retriever}: 
We implement a BM25 retriever \citep{Robertson2009ThePR} with Elasticsearch\footnote{\url{https://www.elastic.co/elasticsearch}}.
The parameters of the BM25 algorithm are set as default values in Elasticsearch ($b=0.75,k_1=1.2$). 
For ReCDS-PAR, we index the title and abstract of a PubMed article as separate fields and the weights given to the two fields when retrieving are empirically set as $3:1$.

\textbf{Dense retriever}:
Dense retrievers represent the patients and articles in a low dimensional space using BERT-based encoders and perform retrieval based on maximum inner-product search.
Concretely, we denote the encoder as $f$, and $\mathbf{e}_d=f(d)$ refers to the low-dimensional embedding generated by the encoder for a given passage $d$. Then for a query patient $q$ and an article $a$ in our retrieval corpus $\mathcal{A}$, the relevance score between them is defined as the inner product of their embeddings: $s_{\text{dense}}(q,a)=\mathbf{e}_q\cdot \mathbf{e}_a$.
The similarity score $s_{\text{dense}}(q,p)$ between $q$ and a patient $p \in \mathcal{P}$ is defined similarly.

We first try direct transferring of Sentence-BERT \citep{Reimers2019SentenceBERTSE} and Contriever \citep{Izacard2021UnsupervisedDI}, two widely-used dense retrievers pre-trained on MS MARCO \citep{Campos2016MSMA}, a general domain retrieval dataset of large scale.
Then we train our own dense retrievers by fine-tuning pre-trained encoders on the PMC-Patients dataset. 
To be specific, for a given query patient $q_i$, a similar patient / relevant article $p_i^+$, and a set of dissimilar patients / irrelevant articles $p_{i,1}^-, p_{i,2}^-,\dots, p_{i,n}^-$ from the training data, we use the negative log-likelihood of the positive passage as the loss function:
$$ L(q_i, p_i^+, p_{i,1}^-,p_{i,2}^-,\dots,p_{i,n}^-) = -\log\frac{e^{s_{\text{dense}}(q_i, p_i^+)}}{e^{s_{\text{dense}}(q_i, p_i^+)} + \sum_{j=1}^ne^{s_{\text{dense}}(q_i, p_{i,j}^-)}} $$
We train the dense retrievers with in-batch negatives \cite{Karpukhin2020DensePR}, where $p_{i,j}^- \in \{ p_{k}^+\ |\ k \neq i \}$.

We train several different encoders, all of which are Transformer encoders \citep{attention} initialized by domain-specific BERT \citep{devlin-etal-2019-bert}, including PubMedBERT \citep{Gu2022DomainSpecificLM}, Clinical BERT \citep{alsentzer2019publicly}, BioLinkBERT \citep{yasunaga2022linkbert}, and SPECTER\citep{Cohan2020SPECTERDR}.
For the ReCDS-PPR task, only one encoder is used, while for the ReCDS-PAR task, we train two independent encoders to encode patients and articles separately, due to their structural differences.

\textbf{Nearest Neighbor (NN) retriever}:
We assume that if two patients are similar, then their respective relevant article and similar patient sets should have a high overlap degree, based on which we implement the following NN retriever similar to \cite{jin2023lader}. 
For each patient in the training queries $p\in\mathcal{P}$, we define its relevant article set as $\mathcal{R}(p) = \{a|a\in\mathcal{A},Rel(p,a)=1 \}$.
For each query patient $q$, we first retrieve top $K$ similar training patients $p_1, p_2, \dots, p_k\in\mathcal{P}$ as its nearest neighbors using BM25\footnote{We also try using fine-tuned dense retrievers which give suboptimal performance.}. 
We take the union of their relevant articles as the candidate set: $$\mathcal{C}(q)=\mathcal{R}(p_1)\cup\mathcal{R}(p_2)\cup\cdots\cup\mathcal{R}(p_K)$$
Then the candidate articles $c_i\in\mathcal{C}(q)$ are ranked by relevance scores $s_{\text{NN}}(q, c_i)$ defined as:
$$ s_{\text{NN}}(q,c_i) = \sum_{k=1}^Ks_{\text{BM25}}(q,p_k)I\{c_i\in \mathcal{R}(p_k)\} $$
NN retriever for ReCDS-PPR is implemented similarly.

\section{Results}\label{result}

In this section, we will first analyze the characteristics of the PMC-Patients dataset in Section \ref{dataset_result}, including basic statistics and patient diversity.
We then show the dataset is of high quality in terms of the summary extraction and the relation annotation in Section \ref{human_eval}.
In the end, we present the performance of baseline methods on the ReCDS-PAR and ReCDS-PPR benchmarks in Section \ref{benchmark_result}.

\subsection{PMC-Patients Dataset} \label{dataset_result}

\textbf{Scale:} Table \ref{tab:ps_basic_stat} shows the basic statistics of patient summaries in PMC-Patients, in comparison to MIMIC, the largest publicly available clinical notes dataset, and TREC CDS, a widely-used dataset for ReCDS.
For MIMIC, we report the statistics of discharge summaries of both MIMIC-\uppercase\expandafter{\romannumeral3} and MIMIC-\uppercase\expandafter{\romannumeral4}.
For TREC CDS, we combine the data released in three years' CDS tracks (2014-2016) and use the ``description'' fields.
PMC-Patients contains $167$k patient summaries extracted from $141$k PMC articles, making
it the largest patient summary dataset in terms of the number of patients, and the second largest in terms of the number of notes. 
Besides, PMC-Patients has $3.1$M patient-article relevance annotations, which is over $27\times$ the size of TREC CDS ($113$k in total).
PMC-Patients also provides the first large-scale patient-similarity annotations, consisting of $293$k similar patient pairs.

\begin{table*}[ht!]
    \small
    \centering
    \begin{tabular}{lcccccc}
    \toprule
    \multirow{2}{*}[-1.5ex]{\textbf{Dataset}} & \multicolumn{2}{c}{\textbf{Count}} &\multirow{2}{*}[-0.3ex]{\textbf{\makecell{Average\\ Length \\ (words)}}} & \multicolumn{2}{c}{\textbf{Relations}}\\
    \cmidrule(r){2-3} \cmidrule(r){5-6}&
     \textbf{Patients} & \textbf{Notes} & & \textbf{\makecell{Relevant \\ Articles}} & \textbf{\makecell{Similar \\ Patients}}\\
    \midrule
    PMC-Patients (ours) & \textbf{167k} & 167k & 410 & \textbf{3.1M} & \textbf{293k} \\
    MIMIC-\uppercase\expandafter{\romannumeral3} (d.s.) & 41k & 60k & 1,282 & -- & --\\
    MIMIC-\uppercase\expandafter{\romannumeral4} (d.s.) & 146k & \textbf{332k} & \textbf{1,480} & -- & --\\
    TREC CDS (all) & 90 & 90 & 92 & 113k & -- \\
    \bottomrule
    \end{tabular}
    \caption{
    Statistics of PMC-Patients, in comparison to MIMIC (d.s.: discharge summaries), and TREC CDS (2014-2016).
    } 
    \label{tab:ps_basic_stat}
\end{table*}

\begin{figure}[ht!]
    \centering
    \includegraphics[width=0.95\linewidth]{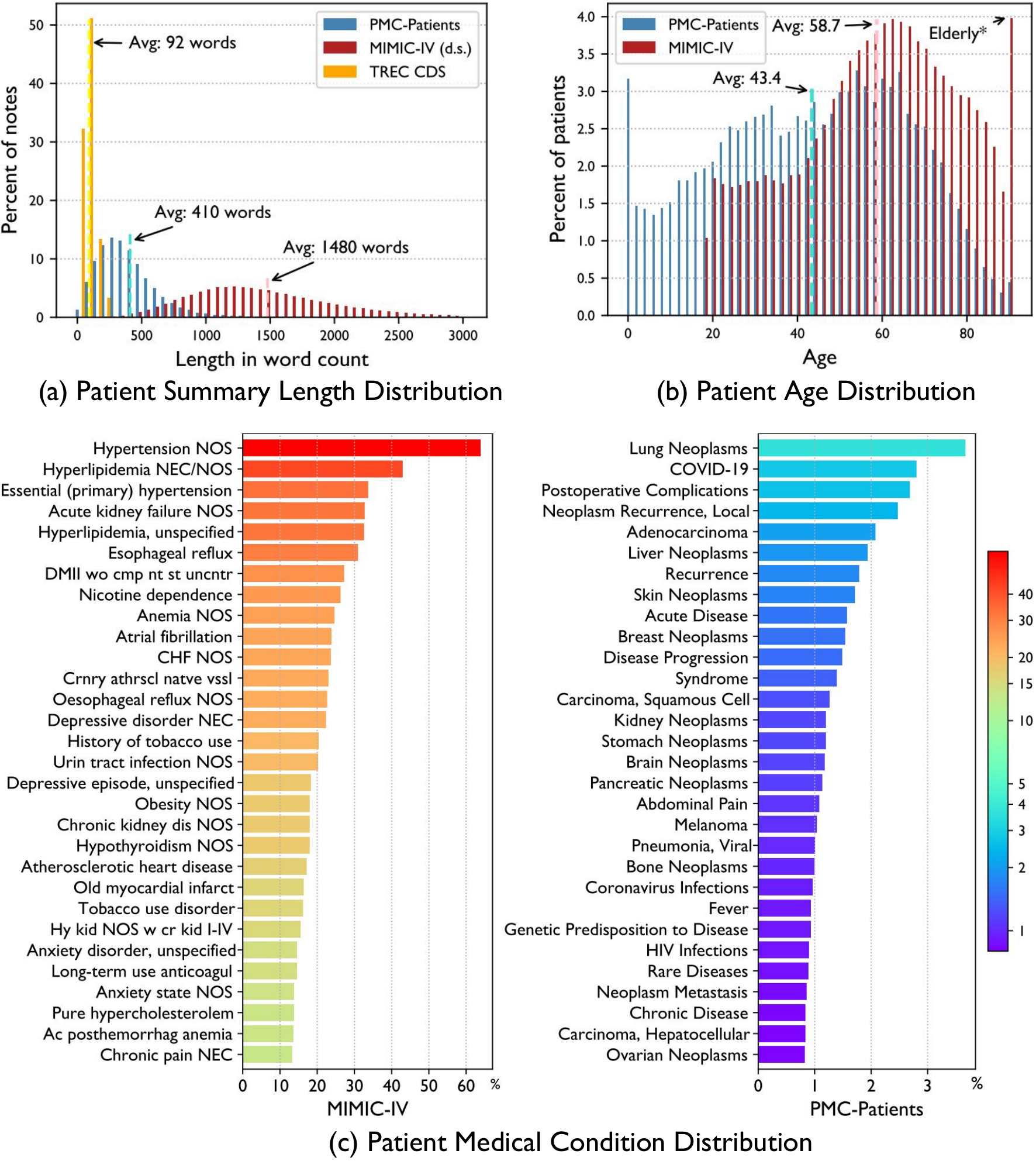}
    \caption{(a): Length distributions of PMC-Patients compared to MIMIC-\uppercase\expandafter{\romannumeral4} discharge summaries and TREC CDS descriptions (x-axis truncated).
    (b): Patient age distributions of PMC-Patients compared to MIMIC-\uppercase\expandafter{\romannumeral4}. *Exact ages of patients older than 89 years old are obscured in MIMIC and thus taken as 90 in the figure.
    (c): Relative frequency of top 30 ICD codes in MIMIC-IV (left) and MeSH Diseases terms in PMC-Patients (right). The colors are associated with relative frequency, and the color bar attached to the figure illustrates this. 
    }
    \label{fig:dataset}
\end{figure}

\textbf{Length:} On average, PMC-Patients summaries are much longer than TREC descriptions ($410$ v.s. $92$ words), but shorter than MIMIC discharge summaries ($410$ v.s. over 1k words). 
Figure \ref{fig:dataset} (a) presents the length distributions of PMC-Patients, TREC CDS descriptions, and MIMIC-\uppercase\expandafter{\romannumeral4} discharge summaries.

\textbf{Demographics:}
The age distributions of PMC-Patients and MIMIC-\uppercase\expandafter{\romannumeral4}
are presented in Figure \ref{fig:dataset} (b). 
There are too few patients to observe the age distribution in TREC CDS, so we don't include it in the figure.
On average, patients in PMC-Patients are younger than MIMIC-\uppercase\expandafter{\romannumeral4} ($43.4$ v.s. $58.7$ years old), and patient ages are more evenly distributed ($6.39$ v.s. $6.09$ Shannon bits). 
PMC-Patients covers pediatric patients while MIMIC-\uppercase\expandafter{\romannumeral4} does not.
The gender distribution in both datasets is balanced. PMC-Patients consists of $52.5$\% male and $47.5$\% female, while MIMIC-\uppercase\expandafter{\romannumeral4} consists of $48.7$\% male and $51.3$\% female.

\textbf{Medical conditions}: We also analyze the medical conditions associated with the patients. For PMC-Patients, we use the MeSH Diseases terms of the articles, and for MIMIC, we use the ICD codes\footnote{There are both ICD-9 and ICD-10 codes in MIMIC-\uppercase\expandafter{\romannumeral4}.}.
The most frequent medical conditions are shown in Figure \ref{fig:dataset} (c). 
In PMC-Patients, the majority of frequent conditions are related to cancer, with the exception of COVID-19 as the second most frequent condition. 
In MIMIC-\uppercase\expandafter{\romannumeral4}, severe non-cancer diseases (e.g. hypertension) have the highest relative frequencies, and their absolute values are much higher than those of the most frequent conditions in PMC-Patients.
For example, hypertension and lung neoplasms are the most frequent condition in MIMIC and PMC-Patients, respectively. Over 60\% of MIMIC patients have hypertension, while less than 4\% of patients in PMC-Patients have lung neoplasms. 
In addition, PMC-Patients covers $4,031/4,933$ (81.7\%) MeSH Diseases terms, relatively more than the $8,955/14,666$ (61.1\%) ICD-9 codes and $16,464/95,109$ (17.3\%) ICD-10 codes covered by MIMIC-\uppercase\expandafter{\romannumeral4}.

\subsection{Dataset Quality Evaluation} \label{human_eval}
\subsubsection{Patient summary extraction}
In this section, we evaluate the quality of the automatically extracted patient summaries and demographics in PMC-Patients.
The evaluation is performed on a random sample of $500$ articles from the benchmark test set.
Two senior M.D. candidates are employed to label the patient note spans at the paragraph level and the patient demographics.
Agreed annotations are directly considered as ground truth, while disagreed annotations are discussed until a final agreement is reached.

Table \ref{tab:ps_human_eval} shows the extraction quality of PMC-Patients and the two human experts against the ground truth. 
A total of $604$ patients are extracted by human experts.
The patient note spans extracted in PMC-Patients are of high quality with a larger than 90\% strict F1 score. 
The extracted demographics are close to 100\% correct. 
Besides, two annotators exhibit a high level of agreement, with most disagreements being minor differences regarding the boundary of a note span.

\begin{table}[ht!]
    \small
    \centering
    \begin{tabular}{lccc}
    \toprule
    \textbf{Quality} & \textbf{Note Span} & \textbf{Age} & \textbf{Gender} \\
    \midrule
    PMC-Patients & 91.24 & 99.77 & {100.0} \\
    Expert A & {97.34} & {100.0} & {100.0} \\
    Expert B & 97.28 & 99.91 & 99.49 \\
    \bottomrule
    \end{tabular}
    \caption{Extraction quality of the PMC-Patients dataset and two experts against the ground truth. Note span recognition is evaluated by F1 score. Age recognition is evaluated by min(annotated\_age, true\_age)/max(annotated\_age, true\_age). Gender recognition is evaluated by accuracy. All numbers are percentages.}
    \label{tab:ps_human_eval}
\end{table}

\subsubsection{Patient-level relation annotation}
To evaluate the quality of patient-level relation annotations in PMC-Patients, we retrieve top $5$ relevant articles and top $5$ similar patients using BM25 for each patient extracted by the human experts in the previous section ($604$ patients from $500$ articles), resulting in over 3k patient-article and 3k patient-patient pairs for human annotation.
To annotate patient-article relevance, we follow the guidelines of the TREC CDS tracks \citep{simpson2014overview,roberts2015overview, roberts2016overview}, where we annotate the type of clinical question that can be answered by an article about a patient, including diagnosis, test, and treatment.
To annotate patient-patient similarity, we follow the recommendations from  \citep{seligson2020recommendations}, where we annotate whether two patients are similar in multiple dimensions: features, outcomes, exposure, and others.
To assess the binary relational annotations in PMC-Patients against the multi-dimensional human annotations, we simply convert the latter into an integer score by counting the number of relevant or similar aspects.
For example, if two patients are annotated as similar in terms of ``features" and ``outcomes", we will give it a score of $2$.

Figure \ref{fig:relation_quality} shows the distributions of the human scores (x-axis) grouped by the relation annotations in PMC-Patients (Irrelevant v.s. Relevant and Dissimilar v.s. Similar). 
T-test shows that patient-article and patient-patient pairs with PMC-Patients annotations have significantly higher human scores than those without ($p<0.01$ for both cases). 
Besides, almost all positive pairs are considered relevant/similar by a human expert, indicating PMC-Patients automatic relational annotations achieve quite high precision.

\subsection{ReCDS Benchmark Results} \label{benchmark_result}

The performance of various baseline methods on the test set of two ReCDS tasks is shown in Table \ref{tab:PPR_PAR_results}. 
Surprisingly, BM25 remains a strong baseline that achieves the best performance on MRR for both tasks and also performs the best on nDCG@10 for ReCDS-PPR.
This indicates the importance of matching the exact words in the case reports for retrieving similar patients or relevant articles.
Sentence-BERT and Contriever, two dense retrievers trained on the general domain MS MARCO dataset, do not generalize well to our ReCDS tasks. 
Their performance on all metrics is much worse than the BM25 baseline, which is consistent with previous studies \citep{Thakur2021BEIRAH, Kim2022ApplicationsAF} that dense retrievers may fail to perform zero-shot retrieval in specific domains such as biomedicine.

On the other hand, dense retrievers fine-tuned on PMC-Patients show significant performance improvements over general domain retrievers, indicating the importance of domain-specific fine-tuning.
Although BM25 performs better on MRR, fine-tuned retrievers achieve the highest P@10 on both tasks and the highest nDCG@10 on ReCDS-PPR.
They also have much higher recall than BM25 which suffers from vocabulary mismatch, showing that semantic matching is indispensable to retrieve more relevant articles or similar patients.
Fine-tuned Clinical BERT performs the worst among other domain-specific BERTs.
This is probably due to the pre-training corpus and tasks of these encoders: PubMedBERT and BioLinkBERT are pre-trained on PubMed; SPECTER and BioLinkBERT incorporate citation graph in pre-training; while Clinical BERT is trained on MIMIC, whose language distribution is quite different from PubMed, and never learns citation relationships.
However, the metrics of the best baseline method are still quite low, highlighting the challenge of the PMC-Patients ReCDS benchmark.

NN retriever generally performs worse than BM25 and dense retrievers, indicating that measuring patient-article relevance based on citation graph distance may not be suitable for the task.

\begin{table}[ht!]
    \small
    \centering
    \resizebox{\linewidth}{!}{
        \begin{tabular}{lcccccccc}
        \toprule
         \multirow{2}{*}[-0.8ex]{\textbf{Method}} & \multicolumn{4}{c}{\textbf{ReCDS-PAR}} & \multicolumn{4}{c}{\textbf{ReCDS-PPR}}\\
         \cmidrule(r){2-5} \cmidrule(r){6-9}
         & \textbf{MRR} & \textbf{Prec} & \textbf{nDCG} & \textbf{Recall} & \textbf{MRR} & \textbf{Prec} & \textbf{nDCG} & \textbf{Recall}\\
        \midrule
        BM25 & \textbf{48.22} & 9.97 & 15.28 & 30.64 & \textbf{22.86} & 4.67 & \textbf{18.29} & 69.66\\
        \thead[l]{Dense retriever\\ (MS MARCO)} \\
        \quad SentenceBERT & 10.58 & 2.71 & 3.53 & 13.52 & 5.28 & 1.17 & 3.88 & 37.55\\
        \quad Contriever & 15.03 & 3.41 & 4.62 & 16.74 & 10.50 & 2.24 & 8.01 & 52.64\\
        \thead[l]{Dense retriever \\ (PMC-Patients)}\\
        \quad PubMedBERT & 42.96 & \textbf{16.08} & 19.51 & \textbf{63.40}& 19.37 & 5.05 & 16.30 & 79.35\\
        \quad Clinical BERT & 24.94 & 8.56 & 10.20 & 48.93 & 10.24 & 2.62 & 7.82 & 67.43\\
        \quad BioLinkBERT & 40.89 & 15.33 & 18.47 & 62.44 & 21.20 & \textbf{5.59} & 18.06 & \textbf{80.49}\\
        \quad SPECTER & 46.41 & 15.59 & \textbf{19.70} & 57.98 & 15.08 & 3.79 & 12.27 & 73.01\\
         NN retriever & 18.76 & 5.93 & 7.03 & 26.55 & 6.30 & 2.40 & 4.83 & 59.49\\
        \bottomrule
        \end{tabular}
    }
    \caption{PAR and PPR performances of baseline retrievers (in percentage). Numbers in bold indicate the best results in each column. 
    Precision (Prec) and nDCG are calculated at 10, and recall is calculated at 1,000.}
    \label{tab:PPR_PAR_results}
\end{table}

\section{Related work}
\subsection{Patient summary dataset and case reports}

Traditionally, patient summary datasets are collected from clinical notes in EHRs, such as MIMIC, MTSamples\footnote{\url{www.mtsamples.com}}, the THYME project \citep{styler2014temporal}, the n2c2\footnote{\url{https://n2c2.dbmi.hms.harvard.edu/}} (originally named i2b2\footnote{\url{https://www.i2b2.org/NLP/DataSets/Main.php}}) project, and the OHNLP Challenges \citep{wang2018overview, wang20202019}. 
However, except for MIMIC, these datasets are limited by size and diversity, typically containing only several hundred to a few thousand clinical note pieces and focusing on specific diseases.

More recently, clinical case reports have been utilized to construct datasets,
but most of the existing works focus on specific tasks such as named entity recognition \citep{schulz2020named, gonzalez2019pharmaconer}, abbreviation resolution \citep{intxaurrondo2018finding}, and semantic similarity \citep{smalheiser2019manual}. They only use case reports as a source of clinical texts, and the resulting datasets are task-oriented, rather than a patient summary dataset.
Only MACCR \citep{caufield2018reference}, CAS \citep{grabar2020cas, grouin2019clinical}, and the E3C project \citep{magnini2021e3c} present patient summary datasets extracted from case reports. Among them, MACCR focuses on curating structured metadata of clinical case reports instead of using free-text patient summaries. CAS and the E3C project mainly focus on European languages such as French and Spanish rather than English, with the dataset scales still limited to several thousand.
In contrast, PMC-Patients is much larger, more diverse, and contains patient-level relation annotations.

\subsection{Retrieval-based clinical decision support}
Due to the lack of an adequate patient summary dataset and the prohibitive costs of manual annotations, there is currently no large-scale ReCDS benchmark dataset available.
Most existing methodology researches on ReCDS-PAR use TREC CDS and TREC Precision Medicine (PM) \citep{roberts2017overview, roberts2018overview,roberts2019overview, roberts2020overview}. 
TREC CDS focuses on retrieving relevant PMC articles for given patient summaries curated by human experts or excerpted from MIMIC with specific intents (e.g. finding treatment/diagnosis).
TREC PM focuses on retrieving relevant literature from PubMed or MEDLINE\footnote{\url{https://www.medline.com/}} and eligible clinical trials that can provide precision medicine-related evidence for a cancer patient, given the patient's cancer type, genetic variants, basic demographics, and other potential factors.
However, each year, only 30-50 patient summaries are released and annotated with patient-article relevance, which also severely limits the patient diversity in these datasets. Furthermore, TREC PM only contains cancer patients.
In contrast, PMC-Patients has a much larger collection of patient summaries (167k) that cover a wider range of medical conditions, and the largest scale of patient-article relevance annotations (3.1M).

To the best of our knowledge, there is no publicly available similar patient retrieval dataset. 
PMC-Patients bypasses the difficulty and expense of patient-level annotations using the PubMed citation graph and construct the first large-scale ReCDS-PPR dataset of $293$k patient-patient similarity annotations.

\section{Discussion}

\subsection{Clinical significance}
ReCDS provides valuable insights for healthcare providers in diagnosis, testing, and treatment of a queried patient, particularly in medically grey zones where high-level evidence is scarce, personalized management for multiple active comorbidities, and off-label use of novel therapeutics.
We here present three case studies in the following section to demonstrate how PMC-Patients can benefit clinicians in different ways.
Specifically, we focus on retrieval of similar patients since this is much less explored than relevant article retrieval.
Table \ref{tab:case_study} shows the three cases under different scenarios with query patient summaries, examples of similar patients retrieved from PMC-Patients, and demonstrations of the clinical significance.
The detailed inputs and outputs for performing case studies are shown in \ref{appendix:case_study}.

\begin{table}[ht!]
    \small
    \centering
    \resizebox{\textwidth}{!} {
    \begin{tabular}{|p{5cm}|p{8cm}|p{6.5cm}|}
    \toprule
    \textbf{Input summary} & \textbf{Retrieval output example} & \textbf{Description and significance} \\
    \midrule
    {\textbf{Patient}: idiopathic thrombocytopenia, glomerulonephritis, and hearing impairment.\qquad \textbf{Scenario}: diagnosis} & {Case Report: Pathogenic MYH9 c.5797delC Mutation in a Patient With Apparent Thrombocytopenia and Nephropathy. (\texttt{patient\_uid}: 8355614-1)} & {Identifying highly-likely combination of associated manifestation and underlying etiology for rare disease like field-experts}\\
    \midrule
    {\textbf{Patient}: history of atrial fibrillation and deep vein thrombosis, signs of cholangitis.\qquad \textbf{Scenario}: test} & {Hemorrhagic cholecystitis causing hemobilia and common bile duct obstruction. (\texttt{patient\_uid}: 6463387-1)} & {Highlighting related active issues for patients with multiple comorbidities thus overcoming cognitive blind-spot}\\
    \midrule
    {\textbf{Patient}: melanoma, initially responsive to BRAF inhibitor but later progressed despite treated with PD-1 inhibitor.\qquad \textbf{Scenario}: treatment} & {Response to Ipilimumab/Nivolumab Rechallenge and BRAF Inhibitor/MEK Inhibitor Rechallenge in a Patient with Advanced Metastatic Melanoma Previously Treated with BRAF Targeted Therapy and Immunotherapy. (\texttt{patient\_uid}: 7334770-1)} & {Out-of-textbook treatment for disease failing standard-of-care, thereby advancing implementation of off-label therapeutics}\\
    \bottomrule
    \end{tabular}
    }
    \caption{Case studies on three patients under different scenarios. For each query patient, we present an example of the retrieved similar patients from PMC-Patients, with corresponding description and significance of assistance in query-patient management.}
    \label{tab:case_study}
\end{table}

The first case involves a diagnostic dilemma of early-onset idiopathic thrombocytopenia, with co-occurred, seemingly unrelated conditions of renal disease, hearing loss, and suspicious family history. 
The top retrieved patient shows \textit{MYH9} mutation \citep{MYH9RDcaseReport}, which is the exact etiology of this case. 
\textit{MYH9}-related thrombocytopenia is extremely rare (1:20,000-25,000) \citep{MYH9RDprevalence} and is thus challenging to diagnose for non-experts.
Other retrieval results also show other possible diagnoses including Alport syndrome \citep{Alport} and anti-basement membrane disease \citep{antiGBM}. 
Its capability to recognize associated features from multiple manifestations and proposing insightful diagnoses is therefore greatly useful, especially in rare diseases.

The second case presents a female patient with a history of atrial fibrillation and deep venous thrombosis who shows acute hepatobiliary symptoms. ReCDS retrieves highly relevant cases, covering most common conditions including cholecystitis \citep{cholecystitis}, bile leak \citep{bileleak}, and Mirizzi syndrome \citep{Mirizzi}.
Impressively, ReCDS is able to bring up potentially dangerous bleeding complications (hemobilia), via suspecting anticoagulation use from her cardiac and thrombotic comorbidities \citep{hemobilia}
This requires further monitoring and testing, thus standing as important reminder in busy clinics where non-major medical problems can be easily ignored. 

The third case asks an open question for treatment of metastatic melanoma failing standard care, pursuing answers in precision medicine similarly as the TREC PM 2020 track \citep{Trec2020PM}. 
The retrieved cases include attempts of ipilimumab/nivolumab rechallenge, BRAFi and MEKi rechallenge \citep{ipinovo_rechallenge}, and single agent PD-1 inhibitor \citep{singlePD1}, each of which providing sound evidence with detailed clinical course background for an oncologist's reference. 
Additionally, the approach itself favors effective treatment combinations (paradoxically thanks to positive report bias), and thus dynamically encourages evidence accumulation towards more promising directions, facilitating future clinical trial designs.

In conclusion, ReCDS can benefit clinicians in various ways, by recognizing rare diseases, overcoming testing blind spots, and advancing treatment evidence. 
With its potential to improve quality of medical care, ReCDS is especially valuable for clinicians in this era of precision medicine and personalized health.

\subsection{Limitations and future work}
Our experiments demonstrate that there is still much room for improvement on the PMC-Patients ReCDS benchmark. 
We outline some potential directions for further research: 
1. Many patient summaries in PMC-Patients have token counts that far exceed BERT's 512 token limit, and truncation is applied in our baselines, which suffer from inevitable information loss. Therefore, retrieval performance may be further enhanced by using efficient transformers \citep{tay2020efficient} such as Big bird \citep{zaheer2020big} and Longformer \citep{beltagy2020longformer}. 
2. Reranking using pre-trained encoders based on cross attentions, including cross-encoders \citep{Nogueira2019MultiStageDR}, poly-encoders \citep{Humeau2020PolyencodersAA}, and ColBERT \citep{Khattab2020ColBERTEA} may significantly improve retrieval performance.
3. Our experiments indicate that both lexical and semantic features are crucial for ReCDS. Previous research has explored combining sparse retrieval and dense retrieval in the general domain \citep{Chen2021SalientPA, Lin2021DensifyingSR}, which may also be useful for the PMC-Patients benchmark.

ReCDS has also been shown helpful in various clinical tasks including question answering \citep{Frisoni2022BioReaderAR}, and patient outcome prediction \citep{Naik2021LiteratureAugmentedCO}, where retrieved relevant articles serve as additional evidence for the model to refer to.
More recently, with the huge success achieved by Large Language Models (LLMs), many studies have explored further augmenting LLMs with retrieval evidence \citep{Guu2020REALMRL, Sachan2021EndtoEndTO, Izacard2022FewshotLW, Ge2023AugmentingZD}.
PMC-Patients can serve both as a benchmark for training and evaluating retrieval systems and as an evidence collection for improving clinical tasks and augmenting clinical LLMs such as ChatDoctor \citep{Yunxiang2023ChatDoctorAM}.

\section{Conclusion}
In this paper, we present PMC-Patients, a large-scale, diverse, and publicly available patient summary dataset with patient-article relevance and patient-patient similarity annotations.
Based on PMC-Patients, we formally define two tasks and provide the largest-scale dataset to benchmark ReCDS: Patient-to-Article Retrieval (ReCDS-PAR) and Patient-to-Patient Retrieval (ReCDS-PPR).
We evaluate various ReCDS systems on PMC-Patients ReCDS benchmarks and show that both tasks are quite challenging, calling for further research.
We also conduct several case studies on our proposed ReCDS benchmark to show the clinical utility of our dataset.

\newpage

\appendix
\section{Patient summary extraction modules}
\label{extraction_module}
\begin{figure}[ht!]
    \centering
    \includegraphics[width=\linewidth]{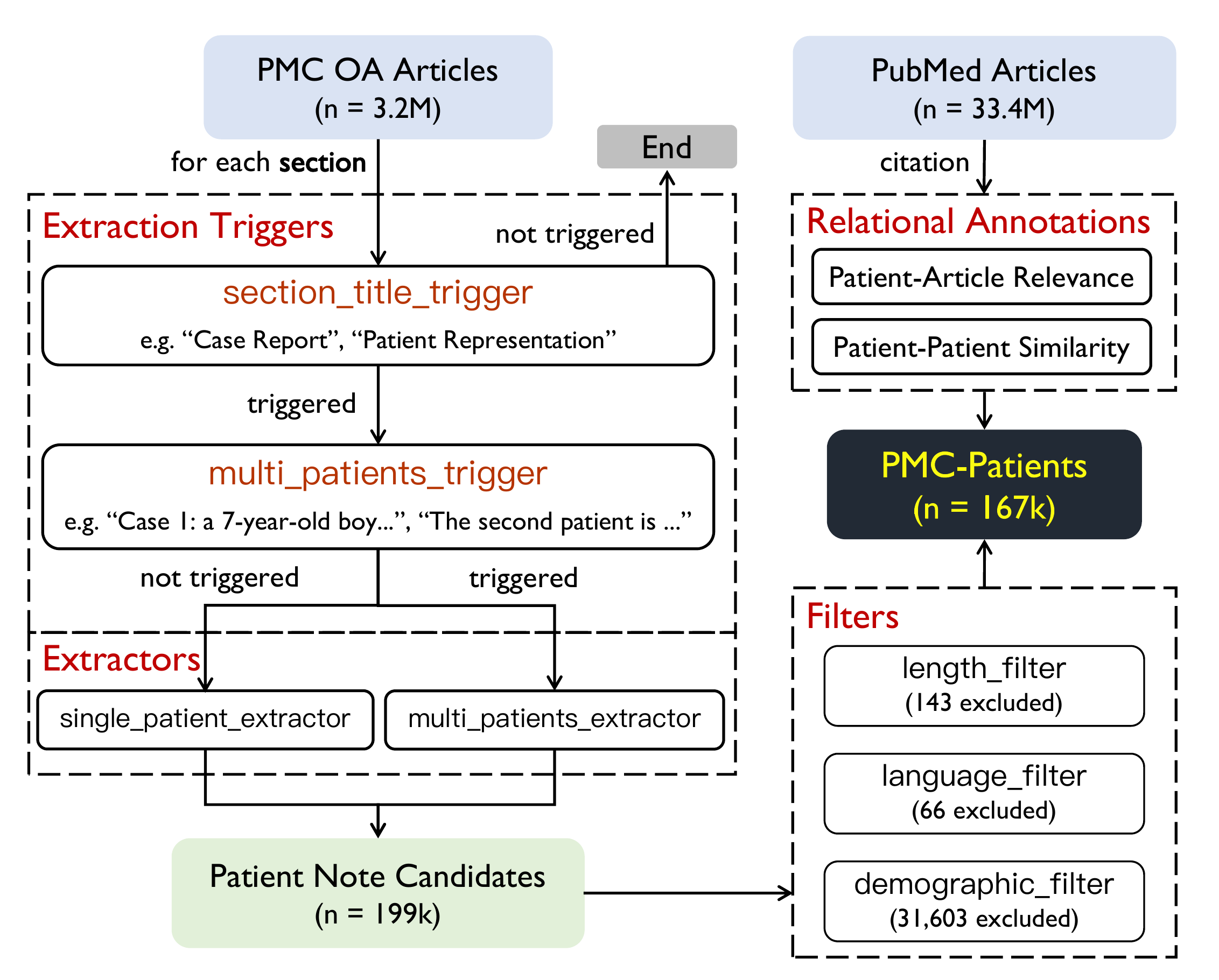}
    \caption{Collection pipeline of PMC-Patients. Patient summaries are identified by \textbf{extraction triggers}, extracted by \textbf{extractors}, and pass various \textbf{filters}. 
    \textbf{Patient-level relations} are annotated using citation relationships in PubMed.}
    \label{fig:pipeline}
\end{figure}

Figure \ref{fig:pipeline} shows the overall collection pipeline of the PMC-Patients dataset.
The detailed implementations of the patient summary extraction modules mentioned in Section \ref{collection} are described below.

\subsection{Extraction triggers}
Extraction triggers are a set of regular expressions to identify whether there are no, one, or multiple potential patient summaries in a given section, basically consisting of two successive triggers:

\texttt{section\_title\_trigger}: Searches in the section title for certain phrases that indicate the presence of patient summaries, such as ``Case Report'' and ``Patient Representation''.

\texttt{multi\_patients\_trigger}: Searches for certain patterns in the first sentence of each paragraph and the titles of the subsections to identify whether multiple patients are presented, such as ``The second patient'' and ``Case 1''.

\subsection{Extractors}
Extractors perform at the paragraph level, i.e. an extracted patient summary always consists of one or several complete paragraphs and no split within a paragraph is performed. 
Depending on whether \texttt{multi\_patients\_trigger} is triggered, different extractors are used:

\texttt{single\_patient\_extractor}: Extracts all paragraphs in the section as one patient summary, if \texttt{multi\_patients\_trigger} is not triggered.

\texttt{multi\_patients\_extractor}: Extracts paragraphs between successive triggering parts (the last one is taken till the end of the section) as multiple patient summaries, if \texttt{multi\_patients\_trigger} is triggered.

\subsection{Filters}
We remove noisy candidates with three filters:

\texttt{length\_filter}: Excludes candidates with less than $10$ words.

\texttt{language\_filter}: Excludes candidates with more than 3\% non-English characters.

\texttt{demographic\_filter}: Identifies the age and gender of a patient using regular expressions and excludes candidates missing either demographic characteristic.
\\

The regular expression rules and parameters in the modules above are generated empirically by manually reading and summarizing hundreds of case reports and then refined on a test set of $100$ articles, which are independent of the $500$ articles used for human evaluation in Section \ref{human_eval}.

\section{Human evaluation of PMC-Patients relation annotations}

 \begin{figure}[ht!]
    \centering
    \includegraphics[width=\linewidth]{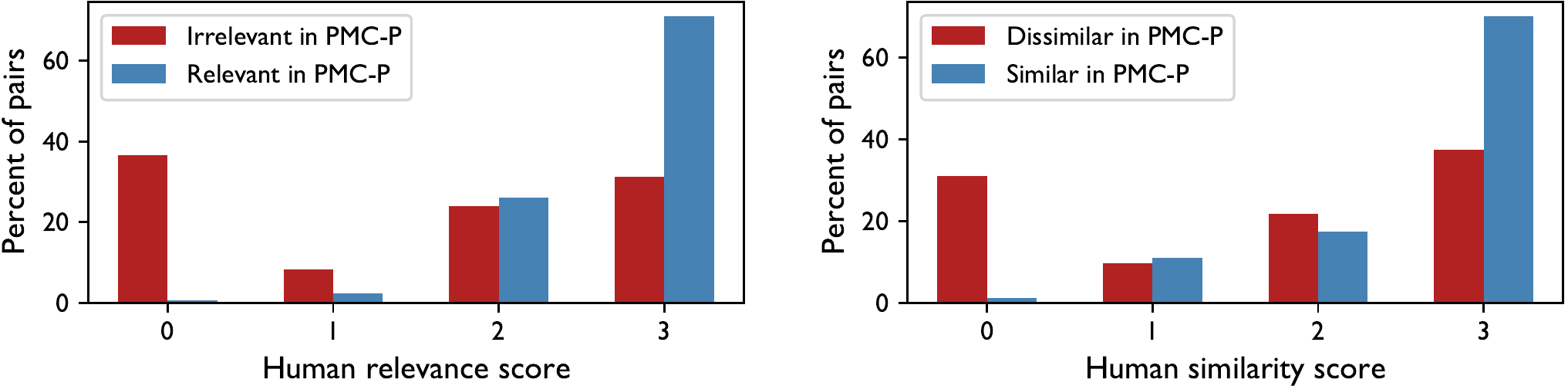}
    \caption{Distributions of the human-annotated relevance (left) and similarity (right) scores grouped by PMC-Patients automatic annotations. 
    }
    \label{fig:relation_quality}
\end{figure}

\section{Top 5 retrieval results in case studies}\label{appendix:case_study}
We show the detailed inputs and retrieved top 5 similar patients and relevant articles for the three case studies in Table \ref{tab:case_study} with the titles of the relevant articles or source articles from which similar patients are extracted.
``Scenario'' is not part of the input since we do not distinguish different scenarios when training the retrievers.
 
\newpage
\subsection{Patient A}

\textbf{Query patient}: A 10-year-old girl presents with cytopenia, hematuria and proteinuria indicating glomerulitis and decreased hearing. Multiple family members also had hearing impairments.

\textbf{Scenario}: Diagnosis.

\textbf{Similar patients}: 
\begin{enumerate}
    \item \texttt{patient\_uid}: 8355614-1.
    
    Case Report: Pathogenic MYH9 c.5797delC Mutation in a Patient With Apparent Thrombocytopenia and Nephropathy.
    \item \texttt{patient\_uid}: 792615-1.

    Atypical anti-glomerular basement membrane disease.
    \item \texttt{patient\_uid}: 7434753-1.

    Athogenic evaluation of synonymous COL4A5 variants in X-linked Alport syndrome using a minigene assay.
    \item \texttt{patient\_uid}: 7752573-1.
    
    A case of hypoparathyroidism, deafness, and renal dysplasia (HDR) syndrome with a novel frameshift variant in GATA3, p.W10Cfs40, lacks kidney malformation.
    \item \texttt{patient\_uid}: 8101677-1.
    
    A Case of Hearing Impairment with Renal Dysfunction.
\end{enumerate}

\textbf{Relevant articles}: 
\begin{enumerate}
    \item \texttt{PMID}: 22567374.
    
    Hearing loss in children with osteogenesis imperfecta.
    \item \texttt{PMID}: 26331839.
    
    Audiometric Characteristics of a Dutch DFNA10 Family With Mid-Frequency Hearing Impairment.
    \item \texttt{PMID}: 16585279.
    
    Universal newborn hearing screening and postnatal hearing loss.
    \item \texttt{PMID}: 18672655.
    
    Hearing rehabilitation in a patient with sudden sensorineural hearing loss in the only hearing ear.
    \item \texttt{PMID}: 16490878.
    
    Idiopathic sudden sensorineural hearing loss in the only hearing ear: patient characteristics and hearing outcome.
\end{enumerate}

\newpage
\subsection{Patient B}

\textbf{Query patient}: A 94 year old female with hx recent PE/DVT, atrial fibrillation, CAD presents with fever and abdominal pain.  An abdominal CT  demonstrates a distended gallbladder with gallstones and biliary obstruction with several CBD stones.

\textbf{Scenario}: Test.

\textbf{Similar patients}: 
\begin{enumerate}
    \item \texttt{patient\_uid}: 3286734-1.
    
    The Successful Treatment of Chronic Cholecystitis with SpyGlass Cholangioscopy-Assisted Gallbladder Drainage and Irrigation through Self-Expandable Metal Stents.

    \item \texttt{patient\_uid}: 6463387-1.
    
    Hemorrhagic cholecystitis causing hemobilia and common bile duct obstruction.

    \item \texttt{patient\_uid}: 4938132-1.
    
    Mirizzi syndrome with an unusual aberrant hepatic duct fistula: a case report.

    \item \texttt{patient\_uid}: 7292700-1.
    
    Biloma: A Rare Manifestation of Spontaneous Bile Leak.

    \item \texttt{patient\_uid}: 7879265-1.
    
    Biliary Peritonitis Caused by Spontaneous Bile Duct Rupture in the Left Triangular Ligament of the Liver after Endoscopic Sphincterotomy for Choledocholithiasis.
\end{enumerate}

\textbf{Relevant articles}: 
\begin{enumerate}
    \item \texttt{PMID}: 24688200.
    
    Fatal abdominal hemorrhage associated with gallbladder perforation due to large gallstones.

    \item \texttt{PMID}: 26265970.
    
    Choledochal Cyst Mimicking Gallbladder with Stones in a Six-Year-Old with Right-sided Abdominal Pain.

    \item \texttt{PMID}: 31938556.
    
    Not everything in the gallbladder is gallstones: an unusual case of biliary ascariasis.

    \item \texttt{PMID}: 22794521.
    
    Spontaneous cholecystocutaneous fistula as a rare complication of gallstones.

    \item \texttt{PMID}: 22028722.
    
    Diagnosis and treatment of multiseptate gallbladder with recurrent abdominal pain.
\end{enumerate}

\newpage
\subsection{Patient C}

\textbf{Query patient}: A 57-year-old man with stage IIIC melanoma was treated with vemurafenib for 8 years with complete response until the disease progressed with brain, lung and liver metastases. After stereotactic radiotherapy, he received nivolumab but progressed again in 2 months later in lung and liver metastases, showing hepatic failure and obstructive jaundice, with LDH value was superior to two-times.

\textbf{Scenario}: Treatment.

\textbf{Similar patients}: 
\begin{enumerate}
    \item \texttt{patient\_uid}: 7662249-1.
    
    Trametinib Induces the Stabilization of a Dual GNAQ p.Gly48Leu- and FGFR4 p.Cys172Gly-Mutated Uveal Melanoma. The Role of Molecular Modelling in Personalized Oncology.

    \item \texttt{patient\_uid}: 7334770-1.
    
    Response to Ipilimumab/Nivolumab Rechallenge and BRAF Inhibitor/MEK Inhibitor Rechallenge in a Patient with Advanced Metastatic Melanoma Previously Treated with BRAF Targeted Therapy and Immunotherapy.

    \item \texttt{patient\_uid}: 6180387-2.
    
    Immune-related adverse events with immune checkpoint inhibitors affecting the skeleton: a seminal case series.

    \item \texttt{patient\_uid}: 5557522-3.
    
    Response to single agent PD-1 inhibitor after progression on previous PD-1/PD-L1 inhibitors: a case series.

    \item \texttt{patient\_uid}: 5109681-1.
    
    Recurrent pleural effusions and cardiac tamponade as possible manifestations of pseudoprogression associated with nivolumab therapy- a report of two cases.
\end{enumerate}

\textbf{Relevant articles}: 
\begin{enumerate}
    \item \texttt{PMID}: 30449777.
    
    Bicytopenia in Primary Lung Melanoma Treated with Nivolumab.

    \item \texttt{PMID}: 23579338.
    
    Vemurafenib and radiation therapy in melanoma brain metastases.

    \item \texttt{PMID}: 33859852.
    
    Fulminant Hepatic Failure after Chemosaturation with Percutaneous Hepatic Perfusion and Nivolumab in a Patient with Metastatic Uveal Melanoma.

    \item \texttt{PMID}: 29434164.
    
    Nivolumab Induces Sustained Liver Injury in a Patient with Malignant Melanoma.

    \item \texttt{PMID}: 26805247.
    
    Multidisciplinary Therapy for Advanced Gastric Cancer with Liver and Brain Metastases.
\end{enumerate}

\bibliographystyle{elsarticle-num}
\bibliography{custom}

\end{document}